\documentclass[runningheads]{llncs}

\usepackage{graphicx} 

\usepackage{kotex}
\usepackage{color}
\usepackage{multirow}
\usepackage{array, booktabs}
\usepackage{arydshln}
\usepackage{calrsfs}
\usepackage{amsmath}
\usepackage{amssymb}

\title{TRACE: Table Reconstruction Aligned to Corner and Edges}
\titlerunning{TRACE}
\author{Youngmin Baek\inst{1,2}\orcidID{0000-0001-7001-4641} \and
Daehyun Nam\inst{3}\orcidID{0009-0001-1927-5739} \and
Jaeheung Surh\inst{4}\orcidID{0009-0007-3296-9085} \and
Seung Shin\inst{3}\orcidID{0009-0008-4653-9614} \and
Seonghyeon Kim\inst{5}\orcidID{0000-0002-8593-1431}}
\authorrunning{Y. Baek et al.}

\institute{
NAVER Cloud, Seongnam-si, Gyeonggi-do, Korea \and
WORKS MOBILE Japan, Shibuya city, Tokyo, Japan \\
\email{youngmin.baek@navercorp.com}  \and
Upstage, Yongin-si, Gyeonggi-do, Korea \\
\email{\{daehyun.nam,seung.shin\}@upstage.ai} \and
Bucketplace, Seocho-gu, Seoul, Korea \\
\email{jh.surh@bucketplace.net} \and
Kakao Brain, Seongnam-si, Gyeonggi-do, Korea \\
\email{	matt.mldev@kakaobrain.com}}

\begin{document}
\maketitle

\begin{abstract}
A table is an object that captures structured and informative content within a document, and recognizing a table in an image is challenging due to the complexity and variety of table layouts. Many previous works typically adopt a two-stage approach; (1) Table detection(TD) localizes the table region in an image and (2) Table Structure Recognition(TSR) identifies row- and column-wise adjacency relations between the cells. The use of a two-stage approach often entails the consequences of error propagation between the modules and raises training and inference inefficiency. In this work, we analyze the natural characteristics of a table, where a table is composed of cells and each cell is made up of borders consisting of edges. We propose a novel method to reconstruct the table in a bottom-up manner. Through a simple process, the proposed method separates cell boundaries from low-level features, such as corners and edges, and localizes table positions by combining the cells. A simple design makes the model easier to train and requires less computation than previous two-stage methods. We achieve state-of-the-art performance on the ICDAR2013 table competition benchmark and Wired Table in the Wild(WTW) dataset.
\end{abstract}

\section{Introduction}

Tables are structured objects that capture informative contents and are commonly found in various documents, such as financial reports, scientific papers, invoices, application forms, etc. With the growth of automated systems, the need to recognize tables in an image has increased. The table recognition task can be divided into two sub-tasks: Table Detection (TD) and Table Structure Recognition (TSR). TD is a task to localize tabular objects in an image, and TSR is a job to identify the adjacency relationship between cells (or contents in cells) within the corresponding table area\cite{gobel2013icdar}.  In the past, the two tasks were conducted using heuristics after extracting the encoded meta-data in the PDF format \cite{rastan2015texus,shigarov2016configurable}. The rule-based approach is out of our scope, and this paper focuses on fully image-based table recognition.

\begin{figure}[t]
  \centering
  \includegraphics*[width=9cm, clip=true]{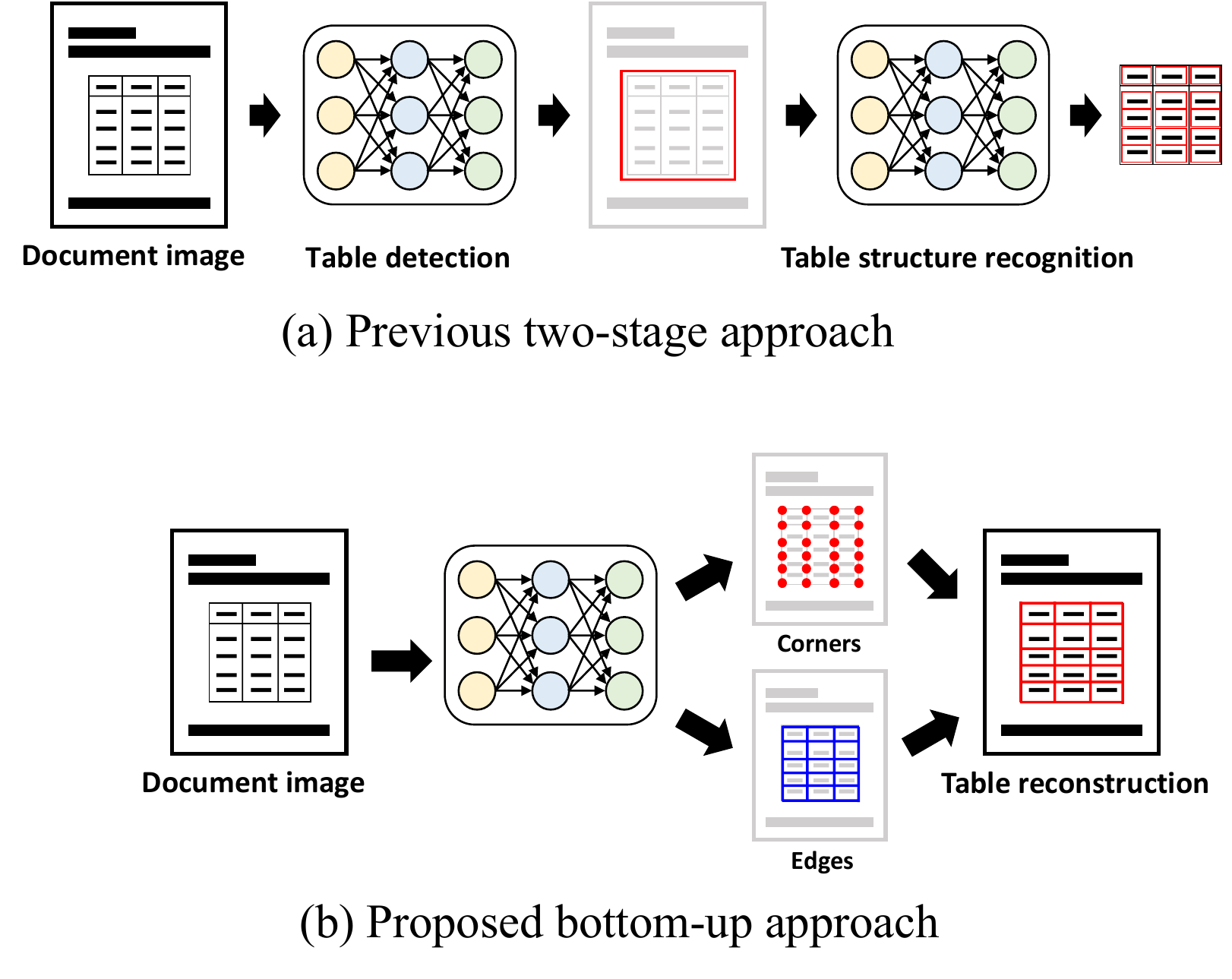}
  \vspace{-5mm}
  \caption{Comparison of our bottom-up approach with previous two-stage ones.}
  \label{fig:overview} 
\end{figure}

Recognizing a table in an image is very challenging because tables have complex layouts and styles. It usually consists of ruling lines, but cells in borderless tables should be semantically separated. Cells and tables have specific considerations as opposed to general objects; there are many empty cells, vast spaces between cells (or contents), extreme aspect ratios, and various sizes. 

With the recent advancement of Deep Learning, TD and TSR performances have been significantly improved~\cite{hashmi2021current}. Most previous methods take the two-stage approach: performing TD and TSR independently, as shown in Fig. \ref{fig:overview}(a). This approach inevitably has two weaknesses. The first is the high training and inference costs. Creating a table recognition system requires more effort since TD and TSR models must be trained separately.
Moreover, the overall table recognition system should serve two models in a row, which causes inefficiencies within the inference process. The second is a constrained end performance bounded by the TD's results. Although TD and TSR are highly correlated and are complementary tasks, there is no interaction between the models to improve each model’s performance. In summary, handling TD and TSR with a single model is highly desirable from a practical point of view. To the best of our knowledge, only two such researches~\cite{paliwal2019tablenet,prasad2020cascadetabnet} have been studied so far.

Toward an end-to-end approach, we analyzed the essential elements constituting tabular objects. A table comprises cells; each cell can be represented using corners and edges. This insight led us to think in the opposite way. Once corners and edges are found, we can reconstruct cells and tables in a bottom-up manner, as depicted in Fig. \ref{fig:overview}(b). Consequently, we propose a novel table reconstruction method called \textit{TRACE (Table Reconstruction Aligned to Corners and Edges)}. In TRACE, a single segmentation model predicts low-level features (corners and edges) rather than the bounding box of cells or contents. After that, simple post-processing can reconstruct the tabular structure.

Using a single model significantly increases the time efficiency in the training and inference phases. We made it possible to reconstruct complex tables by classifying the edges into explicit and implicit lines. Our method shows robust and stable performance on public datasets, including ICDAR2013 and WTW, with state-of-the-art performance.

The main contributions of our paper are summarized as following:
\begin{itemize}
\item We propose a novel end-to-end table reconstruction method with a single model solely from an image.
\item In a bottom-up manner, we propose a table reconstruction method starting from the basic elements of cells, such as corners and edges.
\item The proposed method reconstructs complex tables by classifying edges into explicit and implicit lines.
\item The proposed method achieves superior performance in both clean document datasets (ICDAR13) and natural image benchmarks (WTW).
\end{itemize}

\section{Related Work}

There are two input types for table recognition; PDF and image. The PDF file contains content data, including textual information and coordinates, which are leveraged by the table analysis methods in the early stage \cite{embley2006table}. These rule-based methods rely on visual clues such as text arrangement, lines, and templates. The conventional method only works on  PDF-type inputs. As technology advanced, image-input table recognizers have been proposed mainly based on a statistical machine learning approach as listed in \cite{jorge2006design}. However, it still requires much human effort to design handcrafted features and heuristics. In the era of deep learning, many methods have been studied, showing superior performance compared to the conventional ones \cite{hashmi2021current}. In this paper, we mainly address deep-learning-based related works.

\subsubsection{Table detection(TD)}

TD methods are roughly divided into two categories; object detection-based, and semantic segmentation-based.

The object detection-based approaches adopt state-of-the-art generic object detectors to table detection problems. For example, Faster R-CNN was adopted by \cite{schreiber2017deepdesrt,gilani2017table,siddiqui2018decnt,saha2019graphical,sun2019faster}, and YOLO was used in \cite{huang2019yolo}. Recently, thanks to its instance segmentation ability, Mask R-CNN-based methods \cite{prasad2020cascadetabnet,agarwal2021cdec} were studied. These methods invented techniques such as data augmentation, image transformation, and architecture modification to mitigate a discrepancy between the nature of tables and objects in terms of aspect ratio and sparse visual features.

There were several attempts to apply the semantic segmentation method FCN\cite{long2015fully} on table localization, such as \cite{yang2017learning,he2017multi,kavasidis2019saliency,paliwal2019tablenet}. However, differing from the object segmentation task, the unclear boundary and sparse visual clues in the table region limit the capabilities of these approaches to find accurate table segments.

\subsubsection{Table structure recognition(TSR) } 

The primary purpose of TSR is to identify the structural information of cellular objects. Structure information can be (1) row, column coordinates, and IDs, (2) structural description, (3) connections between contents. According to the problem definition, TSR methods can be categorized into three approaches; detection-based, markup generation-based, and graph-based.

Firstly, many studies have utilized a detection-based approach. However, detection targets are different; row/column regions, cell/content bounding boxes, and separators. Researchers in \cite{schreiber2017deepdesrt,siddiqui2019rethinking,siddiqui2019deeptabstr} have proposed to detect row/column regions based on segmentation and off-the-shelf object detectors. Several studies have proposed detectors for detecting cells or their contents \cite{prasad2020cascadetabnet,raja2020table,qiao2021lgpma,zheng2021global}. In recent years, a split-merge approach has emerged as a popular technique for TSR, in which the separators between cells are initially detected and then subsequently merged  \cite{tensmeyer2019deep,khan2019table,zhang2022split,lin2022tsrformer,ma2023robust}. This strategy is particularly effective in representing the complex layout of tables.

Secondly, markup generation-based approaches try to generate LaTex code or HTML tag directly rather than identifying the coordinates of cellular objects. Collecting LaTex and HTML tags of the table is beneficial for synthetic data generation. Therefore, a large size synthetic table datasets by rendering from the tags have been released such as Table2Latex\cite{deng2019challenges}, TableBank\cite{li2020tablebank}, and  PubTabNet\cite{zhong2020image} In general, the encoder-decoder architecture converts images into markup tags. TableFormer differs from other generation methods in that it decodes not only structural tags but also cell box coordinates \cite{nassar2022tableformer}. However, these approaches need a relatively large dataset and are difficult to handle complex tables in natural scenes.

Lastly, graph-based methods that treat words or cell contents as nodes have been proposed. They analyze the connection between cell relationships using a graph neural network \cite{qasim2019rethinking,chi2019complicated,riba2019table,li2021gfte}. When a node pair is found, it determines whether two nodes are the same row or column and further performs table localization. However, the biggest problem with this method is that it requires a content detection process or additional input from PDF to acquire content. At the same time, it is difficult to deal with empty cells.

\subsubsection{End-to-end table recognition}
A few researchers proposed end-to-end table analyses that include TD and TSR. Most used the two-stage pipeline using two separate models for TD and TSR. DeepDeSRT\cite{schreiber2017deepdesrt} adopted Faster-RCNN\cite{ren2015faster} as a table detector, and SSD\cite{liu2016ssd} as a cell detector. RobusTabNet\cite{ma2023robust} used CornerNet\cite{law2018cornernet} for TD, and proposed line prediction model for TSR. Recently, Zheng et al. proposed Global Table Extractor (GTE)\cite{zheng2021global} that used two separate table and cell detectors based on Faster-RCNN.

We found only a couple of literature using a single model so far. TableNet\cite{paliwal2019tablenet} identifies table and column region. The limitation to separate rows was solved using additional heuristics. CascadeTabNet\cite{prasad2020cascadetabnet} proposed a novel approach to classify table and cell regions simultaneously even though they used a single model.
However, the method still has difficulties handling the blank cell, which needs separated branches for handling bordered and borderless tables after model prediction.


\begin{figure*}[t]
	\begin{center}
  		\includegraphics*[width=12cm, clip=true]{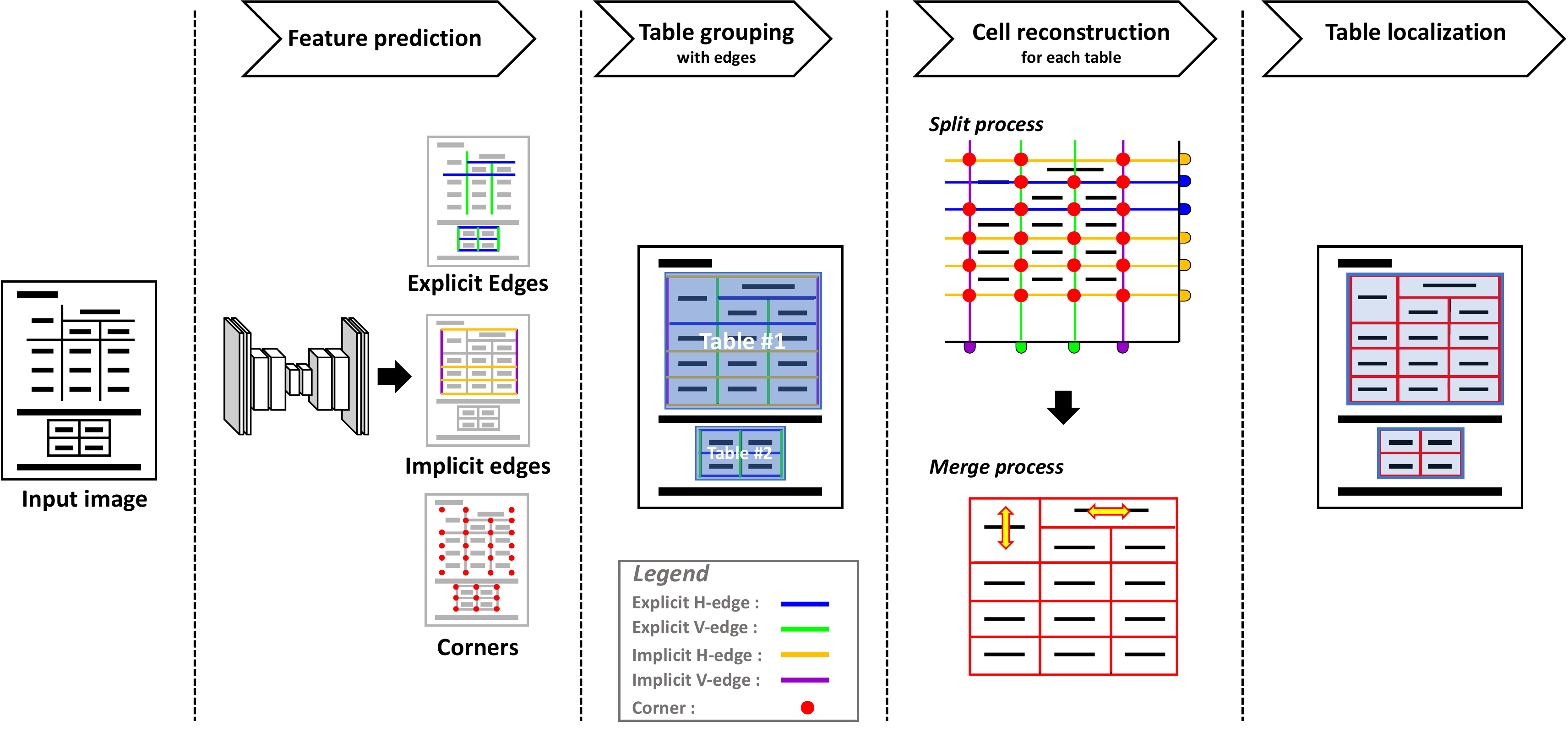}
        \caption{Schematic illustration of our network architecture.}
  	    \label{fig:architecture} 
    \end{center}
\end{figure*} 

\section{Methodology}

\subsection{Overview}


Our method predicts corner and edge segmentation maps that form cells using a deep learning model. The model outputs five segmentation maps; the first channel is used to detect cell corners and the other four channels are used to detect horizontal and vertical edges of a cell box. We define an edge as a separation line between cells, and we predict four types of edges since each edge could represent an explicit or implicit line, either vertically or horizontally. Following the binarization process of the aggregated edge maps, a candidate table region is obtained through connected component labeling. This approach enables the reconstruction of multiple tables with a single inference. In the cell reconstruction step, we simply calculate the position of the separation by projecting horizontal edges to the y-plane and vertical edges to the x-plane for each table candidate region. Here, corners are also used in the search for the separation lines. Then, the spanned cells are merged if there is no edge between the cell bounding boxes. Finally, the table regions are localized through the combination of individual cells. An overview of the pipeline is illustrated in Fig.~\ref{fig:architecture}.
Note that as shown in the legend, the four types of edges are indicated by different colors; explicit horizontal edges in blue, explicit vertical edges in green, implicit horizontal edges in yellow, and implicit vertical edges in purple.


\begin{figure}[t!]
  \centering
  \includegraphics*[width=10.9cm, clip=true]{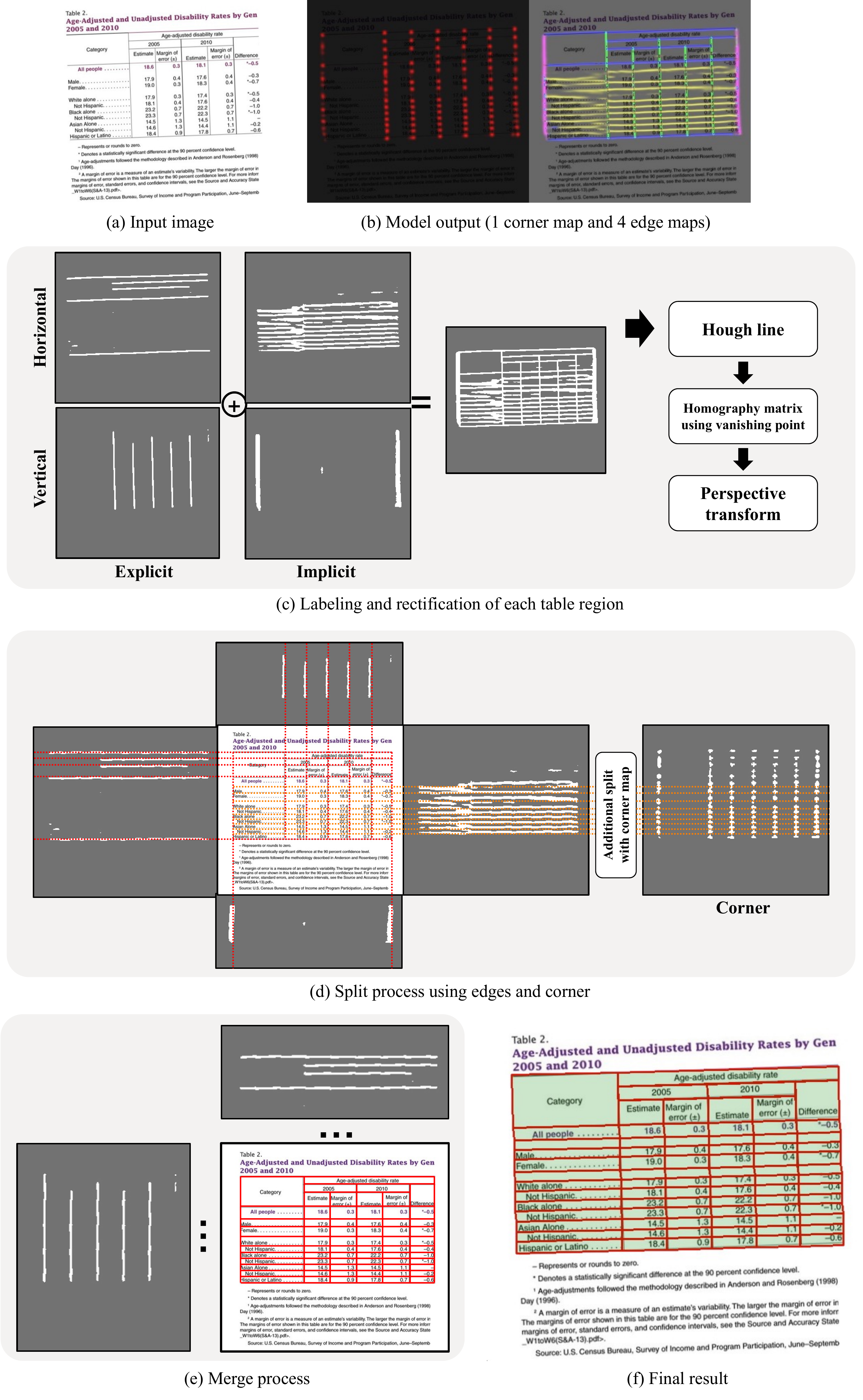}
  \caption{Post-processing of TRACE.}
  \label{fig:postprocessing}
\end{figure}

\subsection{Corner and Edges Prediction Model}

The required corner and edge information are trained using a CNN-based segmentation model. We adopted ResNet-50~\cite{he2016deep} as the feature extraction backbone, and the overall architecture is similar to that of U-Net~\cite{ronneberger2015u}, which aggregates low- and high-level features. The final output has five channels; a corner map, explicit horizontal/vertical edge maps, and implicit horizontal/vertical edge maps.

For ground truth label generation, we need the cell bounding boxes and properties of each edge. For the corner map, we render a fixed sized Gaussian heatmap centered on every corner point of the cell. For the edge map, a line segment is drawn with a fixed thickness on every side of the cell. Here, horizontal- and vertical- edge ground truths are generated on the different channels. Also, a property of the edge indicates whether the line segment is visible or not. If visible, we use the \textit{Explicit Edge} channels, if not, the \textit{Implicit Edge} channels are used.

We use the MSE loss for the objective $L$, defined as,
\begin{equation} \label{eq:weighted_loss}
\begin{aligned}
L & = \sum_{p} \sum_{i} ||S_{i}(p)-S^{*}_{i}(p)||^{2}_{2},
\end{aligned}
\end{equation}

\noindent where $S_{i}(p)$ denotes the ground truth of $i$-th segmentation map, and $S_{i}^{*}(p)$ denote the predicted segmentation map at the pixel $p$.


Note that the TRACE method has two notable features that set it apart from traditional cell detection methods relying on the off-the-shelf object detectors. Firstly, TRACE detects low-level visual features rather than high-level semantic elements such as cell bounding boxes or content bounding boxes. This approach allows for easier learning due to the distinctiveness of low-level visual cues and enables the effective handling of empty cells, which is challenging for other methods. Secondly, TRACE is capable of identifying both explicit and implicit edges when generating separators. The distinction between visible and non-visible lines helps to find separators from the table image. The table reconstruction is achieved through a series of heuristic techniques in the post-processing after the low-level features have been obtained.

\subsection{Data preparation}

As explained in the previous section, we need cell bounding box data with attributes. Unfortunately, public datasets only have two types of bounding box annotation; content bounding box~\cite{gobel2013icdar,gao2019icdar,chi2019complicated} and wired cell bounding box~\cite{long2021parsing}. Some public datasets for structure recognition do not provide any coordinates for content or cell, but structure markup such as HTML tags or LaTex symbols~\cite{li2020tablebank,zhong2020image}. This lack of available data for our approach motivated us to collect our own dataset.

We collected document images from the web including invoices and commonly used goverment documents. Also, some came from TableBank~\cite{li2020tablebank}.  The dataset includes complicated tables with visible and invisible separation lines. The annotators were asked to annotate line segments and its attributes for us to create cell bounding boxes with properties.


It is easy to maintain data consistency of explicit edges since the visual cues are clear. However, data inconsistency issue arise when dealing with implicit edges because it is not clear where to form a line between the cells. To alleviate this issue, the annotators were guided to construct an equidistant bisector line that equally separates the cells.



\subsection{Post-processing for reconstruction}

The procedure for table reconstruction by TRACE is illustrated in Fig. \ref{fig:postprocessing}. Given an input image, the TRACE model predicts corner and edge maps, which are then processed in the following steps.

In the image rectification step, we apply binarization to the inferred segmentation maps, and then use the combination of all binary edge maps to approximate the location of each table. The Hough line transform is used to detect lines in the binary edges. We rectify the table image by making horizontal and vertical lines perpendicular. This rectification is not necessary for most scanned document images.

\begin{figure}[h!]
  \centering
  \includegraphics*[width=12cm, clip=true]{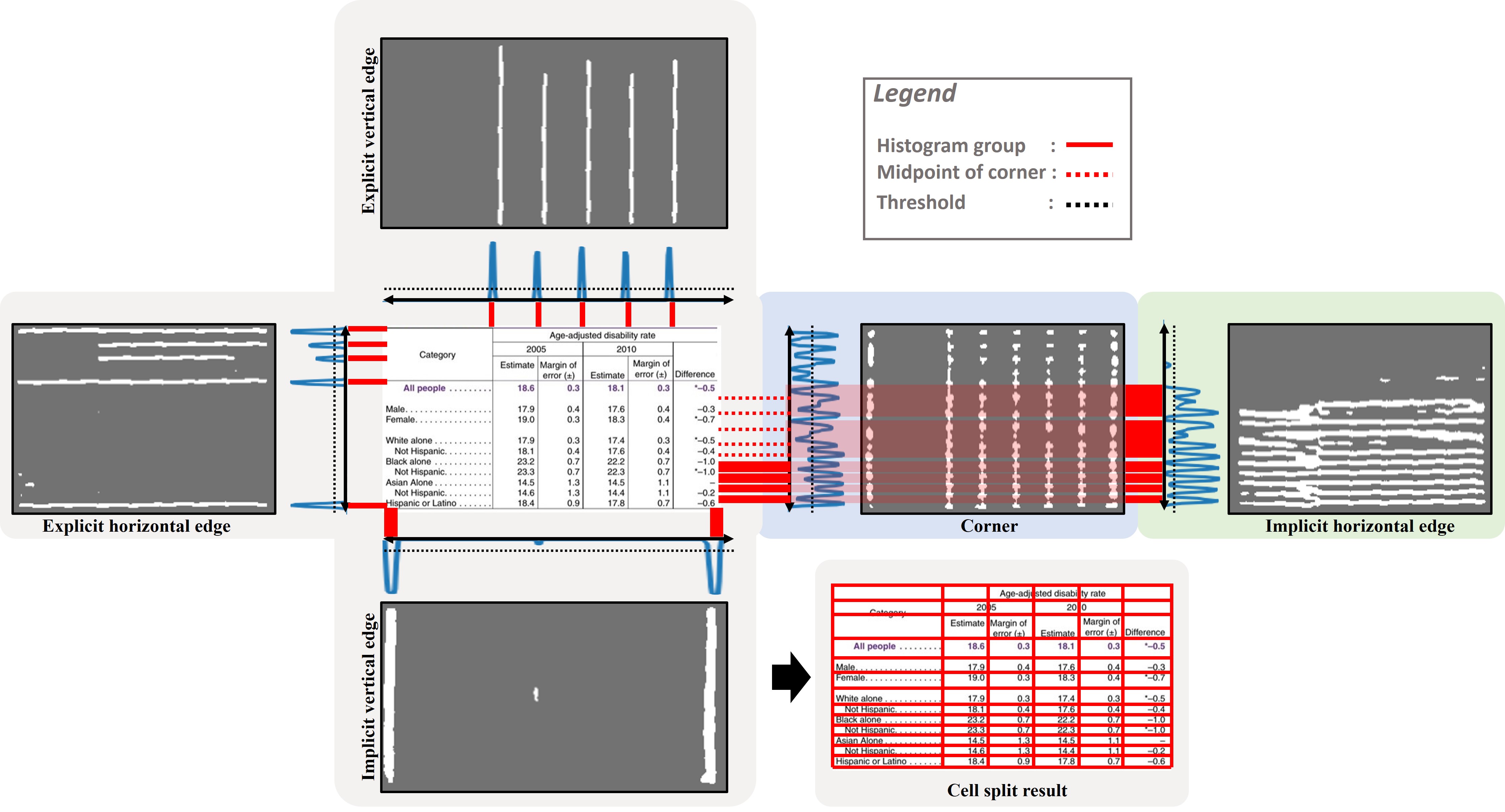}
  \caption{The split process in TRACE is performed to separate the cells of a table. After the binarization of the edge maps, the mid-point of the grouped binary edge maps is determined as the separator. This process is repeated for both the horizontal and vertical axes. To address the issue of ambiguity in implicit horizontal lines, the peak positions of the projected corner map are also utilized.}
  \label{fig:split_process}
\end{figure}

TRACE employs a split-merge strategy, inspired by SPLERGE~\cite{tensmeyer2019deep}, to reconstruct tables. The split process is demonstrated in the Fig. \ref{fig:split_process}. It begins by projecting explicit horizontal edge maps onto the y-plane and explicit vertical edge maps onto the x-plane. The midpoint of the projected edge group on the plane is considered to be the location of each separation line.

In the case of implicit horizontal lines, the projection may not clearly separate them. When a row in the table includes empty cells, the implicit horizontal lines may be combined due to the ambiguity of the content borders. To resolve this issue, we use the corner map. We apply the same binarization and projection processes to the corner map. If the projected line group is thick, with more than two peak points of corners, the final split point is calculated based on the peak points of the corner groups, rather than the midpoint of the line group.





After the split process, the cell merge process is initiated. This process utilizes both explicit and implicit edge maps. The basic rule applied in this process is that if two adjacent cells lacked a binary edge map in the midpoint of their separator, the separator is removed, and the cells are merged. 

Finally, the quadrilateral of the table location is determined by computing the coordinates of the top-left, top-right, bottom-left, and bottom-right corners of all detected cells. If the table image is transformed due to the rectification process, the cell coordinates and table coordinate are unwarped to the image coordinate system.

It is important to note that while the basic post-processing for TRACE is relatively straightforward, the requirements may vary depending on the dataset. For instance, the WTW dataset does not provide annotations for implicit lines, but some images in the dataset contain inner tables with invisible edges. Thus, heuristics may need to be designed based on the end task. This will be discussed in the experiment section, where examples will be presented.

\section{Experiments}

We conducted experiments on public benchmarks, including the ICDAR2013 table competition dataset and Wired Table in the Wild (WTW) dataset, to validate the proposed method.

\subsection{Dataset and evaluation}

To train TRACE, we need cell bounding box data with visibility flags. However, there is no public dataset that satisfies this condition. We collected document images from the web, and manually annotated tables. Our in-house dataset mainly consists of financial and scientific documents. The number of total images is 9717, which are divided into 7783 training images, 971 validation ones, and 963 testing ones. We will soon make a portion of our in-house dataset available to the public. The dataset will include an adjacency relation-based evaluation metric for assessing performance.

ICDAR2013 Table Competition benchmark~\cite{gobel2013icdar} is the table dataset that is commonly evaluated. The dataset is composed of 156 tables in PDF format from the EU/US government website. We only tested our method on this dataset without finetuning. We used the official evaluation protocol. For table detection, we calculated the character-level recall, precision and F1-score along with \textit{Purity} and \textit{Completeness}. \textit{Purity} increases if a detected table does not include any character that are not also in the GT region, \textit{Completeness} counts whether a table includes all characters in the GT region. For the table structure recognition evaluation, we used adjacency relations-based recall precision measures.

WTW dataset~\cite{long2021parsing} contains not only document images but also images of scenes from the wild. Therefore, it has a variety of tables in terms of types, layouts, and distortions. There are 10,970 training images and 3,611 testing images. WTW only focuses on wired tables. We follow the evaluation metric by the authors; (1) for cell detection, cell box-level recall/precision with IoU=0.9 are used, (2) for structure recognition, cell adjacency relationship-level recall/precision from the matched tabular cells with IoU=0.6 are used.

Our method is unable to evaluate widely-used table datasets for tag generation, such as SciTsr\cite{chi2019complicated} and TableBank\cite{li2020tablebank} since evaluation metrics cell content with text attributes. This necessitates the use of additional OCR results, which beyond the scope of image-based table reconstruction.




\subsection{Implementation detail}

In the training process, we used a ResNet50 backbone pretrained on ImageNet. The longer side of the training images are resized to 1280 while preserving the aspect ratio. The initial learning rate is set to 1e-4, and decayed at every 10k iteration steps. We train the model up to 100k iterations with a batch size of 12. The basic deep learning techniques such as \textit{ADAM}~\cite{kingma2014adam} optimizer, \textit{On-line Hard Negative Mining}~\cite{shrivastava2016training} and data augmentations including color variations, random rotations, and cropping are applied.

In our methodology, the parameters for post-processing were determined empirically. Specifically, binary thresholds were set to 0.5 for explicit edge maps and 0.2 for implicit edge maps. And, if the length of a vertical or horizontal edge was less than 25\% of the corresponding table height or width, the edge was deemed not suitable for the split process and was discarded.

\subsection{Result on Document Dataset(ICDAR2013)}

\begin{table*}[ht!]
\caption{Table Detection Results on ICDAR13 dataset.}
  \centering
  \begin{tabular}{c||c||c|c|c||c|c}
    \hline 
    \textbf{Method} & \textbf{Input type} &  \textbf{Recall} & \textbf{Precision} & \textbf{F1} & \textbf{Complete} & \textbf{Pure} \\
    \hline
    \hline
    Nurminen~\cite{gobel2013icdar}   & PDF   & 90.77 & 92.1  & 91.43 & 114   & 151 \\
    Silva~\cite{silva2010parts}      & PDF   & 98.32 & 92.92 & 95.54 & 149   & 137 \\
    \hdashline
    TableNet~\cite{paliwal2019tablenet}  & Image & 95.01 & 95.47 & 95.47 & - & - \\
    Tran et al.~\cite{tran2015table} & Image & 96.36 & 95.21 & 95.78 & 147   & 141 \\
    TableBank~\cite{li2020tablebank} & Image & -     & -     & 96.25 & -     & - \\
    DeepDeSRT~\cite{schreiber2017deepdesrt} & Image  & 96.15 & 97.40 & 96.77 & - & -\\ 
    GTE~\cite{zheng2021global}       & Image & \textbf{99.77} & \textbf{98.97} & \textbf{99.31} & 147   & 146 \\
    \hline
    Ours~(TRACE) & Image & 98.08 & 97.67 & 97.53 & \textbf{150}  & \textbf{147} \\
    \hline
  \end{tabular}
  \label{tab:result_det_icdar13}
\end{table*}

The table detection results of various methods on ICDAR2013 benchmark are listed in Table.~\ref{tab:result_det_icdar13}. TRACE's F1-measure shows competitive performance with GTE~\cite{zheng2021global}, and it achieved a higher score in terms of \textit{Purity} and \textit{Completeness}.

Some TD methods are not directly compared here, because 1) CascadeTabNet~\cite{prasad2020cascadetabnet} used a subset of test images for evaluation, and 2) CDeC-Net~\cite{agarwal2021cdec} and other methods~\cite{saha2019graphical,huang2019yolo,siddiqui2018decnt} reported their performance using IoU-based metrics. The results of PDF-based methods are only for reference since they detect table regions from PDF-metadata, not from images.

\begin{table*}[ht!]
  \caption{Table Structure Recognition Results on ICDAR13 dataset.}
  \centering
  \begin{tabular}{c||c||c|c|c||c|c|c}
    \hline 
    \textbf{Method} & \textbf{Venue} &  \textbf{Recall} & \textbf{Precision} & \textbf{F1} & \textbf{Approach} & \textbf{GT} & \textbf{PDF} \\
    \hline
    \hline
    Nurminen   & ICDAR13   & 94.09 & 95.12 & 94.60 & TSR   & \checkmark  & \checkmark \\ 
    TabStructNet& ECCV20    & 89.70 & 91.50 & 90.60 & TSR   & \checkmark  & - \\
    SPLERGE & ICDAR19   & 90.44 & 91.36 & 90.89 & TSR   & \checkmark  & - \\
    Split-PDF+Heuristics & ICDAR19   & 94.64 & 95.89 & 95.26 & TSR   & \checkmark  & \checkmark \\
    GTE       & WACV21    & 92.72 & 94.41 & 93.50 & TD+TSR & - & - \\
    GTE(with GT)& WACV21  & 95.77 & 96.76 & 96.24 & TSR   & \checkmark & - \\
    LGPMA       & ICDAR21   & 93.00 & 97.70 & 95.30 & TSR   & \checkmark & - \\
    \hline
    Ours(TRACE)                     & -         & \textbf{96.69} & \textbf{98.47} & \textbf{97.46} & E2E   & - & - \\
    \hline
  \end{tabular}
  \label{tab:result_str_icdar13}
\end{table*}

For the table structure recognition task, TRACE achieved the highest score when comparing with previous works as shown in Table~\ref{tab:result_str_icdar13}. The important point we want to emphasize is that TRACE is the only end-to-end approach. TSR-only methods require the cropped table region in the image, but ours does not. Performing both TD and TSR tasks simultaneously is difficult since the inaccurate table detection results in lowering the end performance. For example, when comparing the result with and without table detection in GTE, the performance dropped by 2.7\%. Our bottom-up approach proved its robustness both in TD and TSR tasks.

As with the TD results, some of previous TSR methods cannot be listed in the result table directly because the authors randomly chose 34 images for testing in order to overcome the lack of training images (e.g. \cite{schreiber2017deepdesrt}, \cite{paliwal2019tablenet})

\subsection{Result on Wild Scene Dataset (WTW)}

\begin{figure}[h!]
  \centering
  \includegraphics*[width=10cm, clip=true]{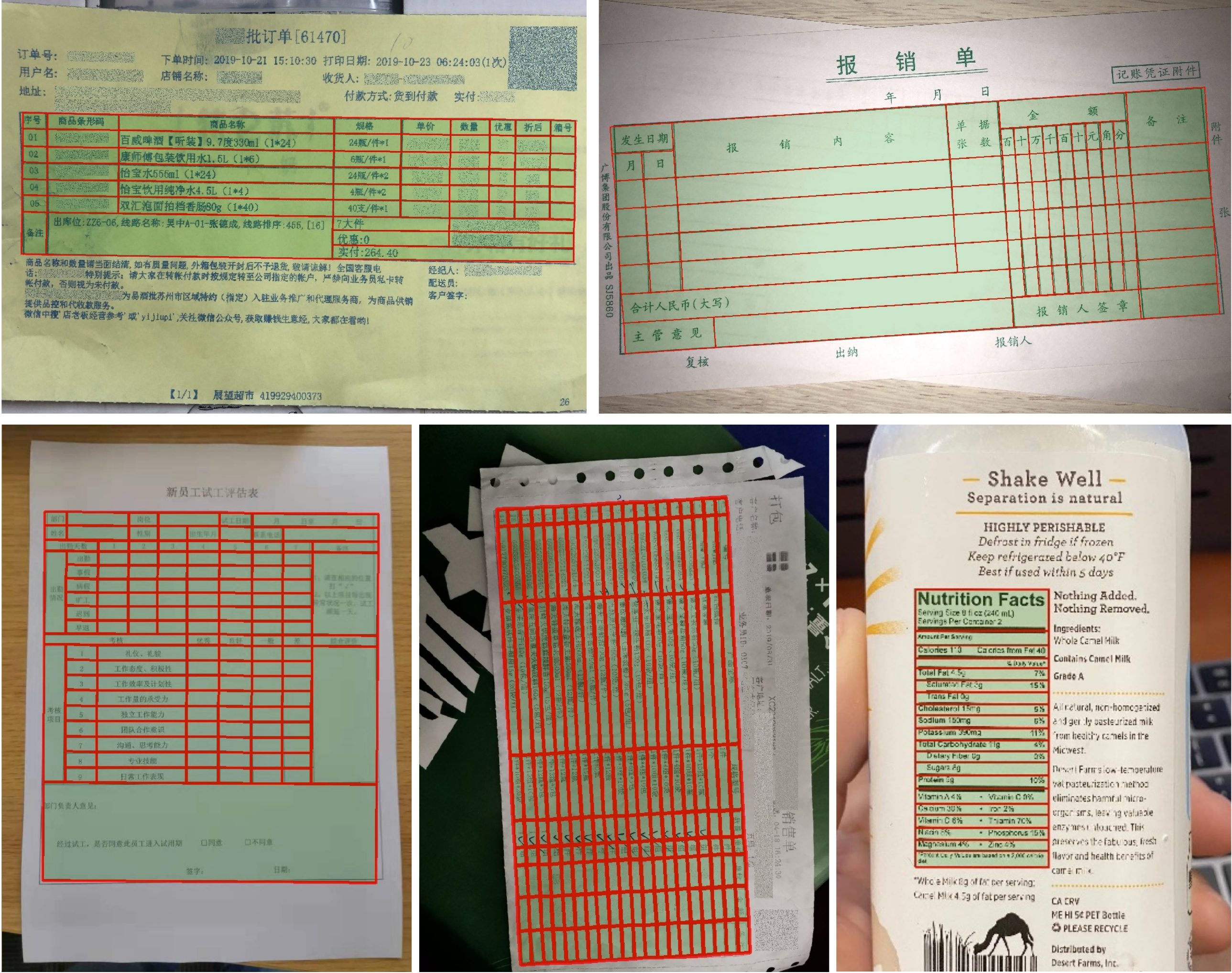}
  \caption{Table reconstruction results on WTW datasets.} 
  \label{fig:result_wtw}
\end{figure}

The WTW dataset only contains wired tables, so, TRACE was trained on this dataset without incorporating implicit edge maps. For the experiment on the WTW dataset, no additional data was utilized for the training process. The results on WTW are visualized in Fig. \ref{fig:result_wtw}.

\begin{table*}[h!]
  \caption{Cell-level detection and adjacency relation-level structure recognition results on the WTW dataset.}
  \centering
  \begin{tabular}{c||c|c|c||c|c|c||c|c}
    \hline 
    \multirow{2}{*}{\textbf{Method}} & \multicolumn{3}{c||}{Physical Coordinates} & \multicolumn{3}{c||}{Adjacency Relation} & \multirow{2}{*}{\textbf{Approach}} & \multirow{2}{*}{\textbf{GT Table}} \\
    \cline{2-7}
     & \textbf{R} & \textbf{P} & \textbf{F1} & \textbf{R} & \textbf{P} & \textbf{F1} & \\
    \hline
    \hline
    Cycle-Centernet~\cite{long2021parsing} & \textbf{78.5} & \textbf{78.0} & \textbf{78.3} & 91.5 & 93.3 & 92.4 & TD+TSR & - \\
    TSRFormer~\cite{lin2022tsrformer} & - & - & - & 93.2 & 93.7 & 93.4 & TSR & \checkmark \\
    NCGM~\cite{liu2022neural}& - & - & - & \textbf{94.6} & 93.7 & 94.1 & TSR & \checkmark \\
    \hline
    Ours~(TRACE) & 63.8 & 65.7 & 64.8 & 93.5 & \textbf{95.5} & \textbf{94.5}& E2E & - \\
    \hline
  \end{tabular}
  \label{tab:result_wtw}
\end{table*}

Cell-level evaluation is conducted on WTW instead of table-level detection. The results on WTW is shown in Table.~\ref{tab:result_wtw}. TRACE achieved the best performances in the structure recognition task. It is also noted that we compared the TSR methods that require GT table regions. TRACE is the result obtained by performing the task end-to-end without table cropping.

We evaluated our method using official parameters provided by the authors of the WTW dataset. Here, the IoU threshold was set to 0.9 for cell detection evaluation. However, we found false positive cases even for sufficiently reasonable results, which means 0.9 is too strict parameter for matching of small cells. Also, some imprecise annotations were found, which is critical for small cell boxes. We estimate that 0.7 is a more appropriate parameter for IoU threshold based on the quantitative analysis. For reference, further experiments with various IoU were conducted as shown in Table.~\ref{tab:result_wtw_iou}.

\begin{table}[h!]
  \caption{Cell detection result with various IoU thresholds for matching cells on WTW dataset. }
  \centering
  \begin{tabular}{c||c|c|c||c}
    \hline 
    \textbf{IoU} & \textbf{R} & \textbf{P} & \textbf{F1} & \textbf{Ref} \\ 
    \hline
    \hline
    0.9 & 63.8  &	65.7 & 	64.8 & Parameter for detection\\
    0.8 & 89.5  &	92.1 &	90.8 & - \\
    0.7 & 93.6	&   96.3 &  94.9 & -\\
    0.6 & 94.8  &	97.5 &	96.1 & Parameter for TSR\\
    \hline
  \end{tabular}
  \label{tab:result_wtw_iou}
\end{table}

\begin{figure*}[h!]
  \centering
  \includegraphics*[width=\textwidth, clip=true]{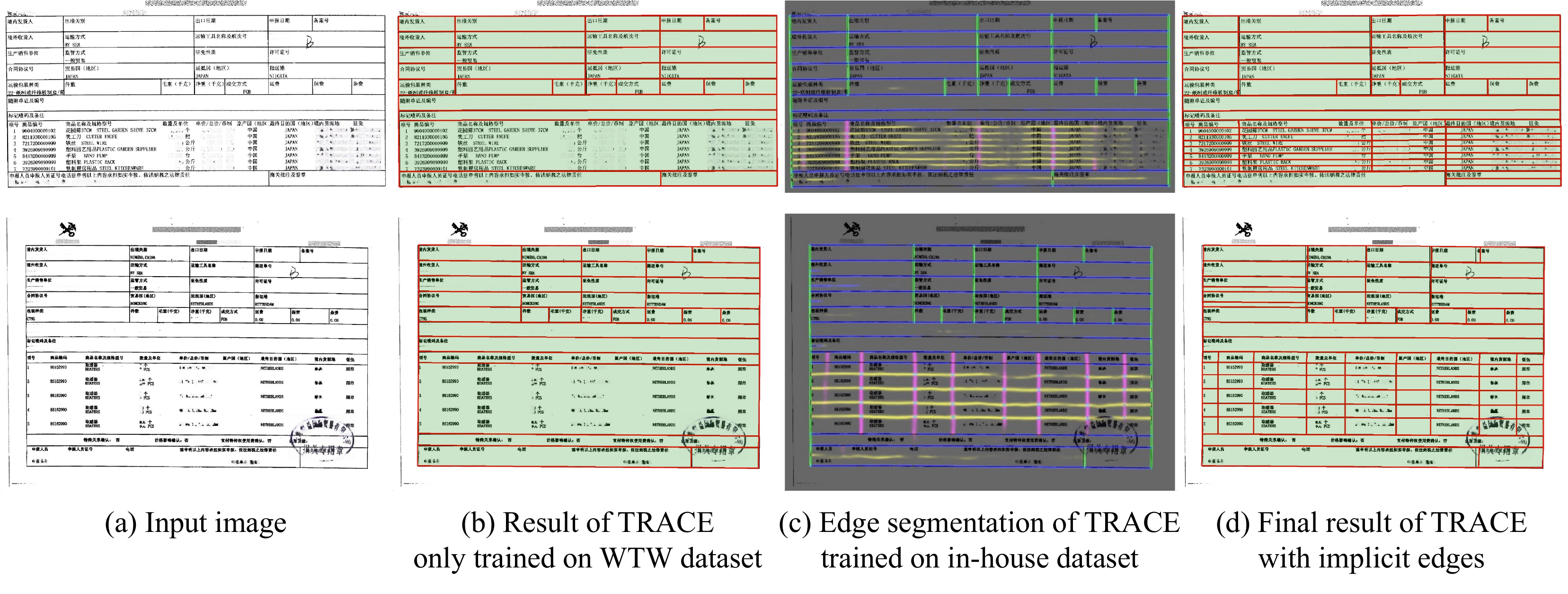}
  \caption{Comparison of TRACE results with and without implicit edge. Even complex documents like custom declarations are correctly recognized with TRACE. } 
  \label{fig:result_implicit_wtw}
\end{figure*}

Figure.~\ref{fig:result_implicit_wtw} visualizes the table reconstruction result of TRACE on WTW dataset. Most of the images in WTW contain wired tables. Yet, there are document images including borderless tables in which product items are listed. The user's expectation is to reconstruct cellular objects that are semantically listed up. Here, we additionally investigate table reconstruction results on customs declarations through TRACE trained on our in-house dataset. The final result was qualitatively better than the human annotated ground truth. By showing correct table reconstruction results on these mixed-type tables, TRACE proved its high usability and generalization ability.

\subsection{Discussion}

\subsubsection{Bordered and borderless table comparison} We separate ICDAR2013 benchmark into bordered and borderless tables, and evaluated them separately. We investigated this to check the difference in difficulty although the number of files differ: 30 bordered and 31 borderless sheets. 
In the TSR task, the F1-score on bordered and borderless tables are 99.4\% and 93.85\%, respectively. Like this, the score on bordered tables are quite high, and the main reason for the performance drop is with the borderless cases. 
To improve this, implicit edges could be counted as separation line candidates, and additional heuristics such as the existence of content can be applied further.



\begin{figure}[h!]
  \centering
  \includegraphics*[width=9cm, clip=true]{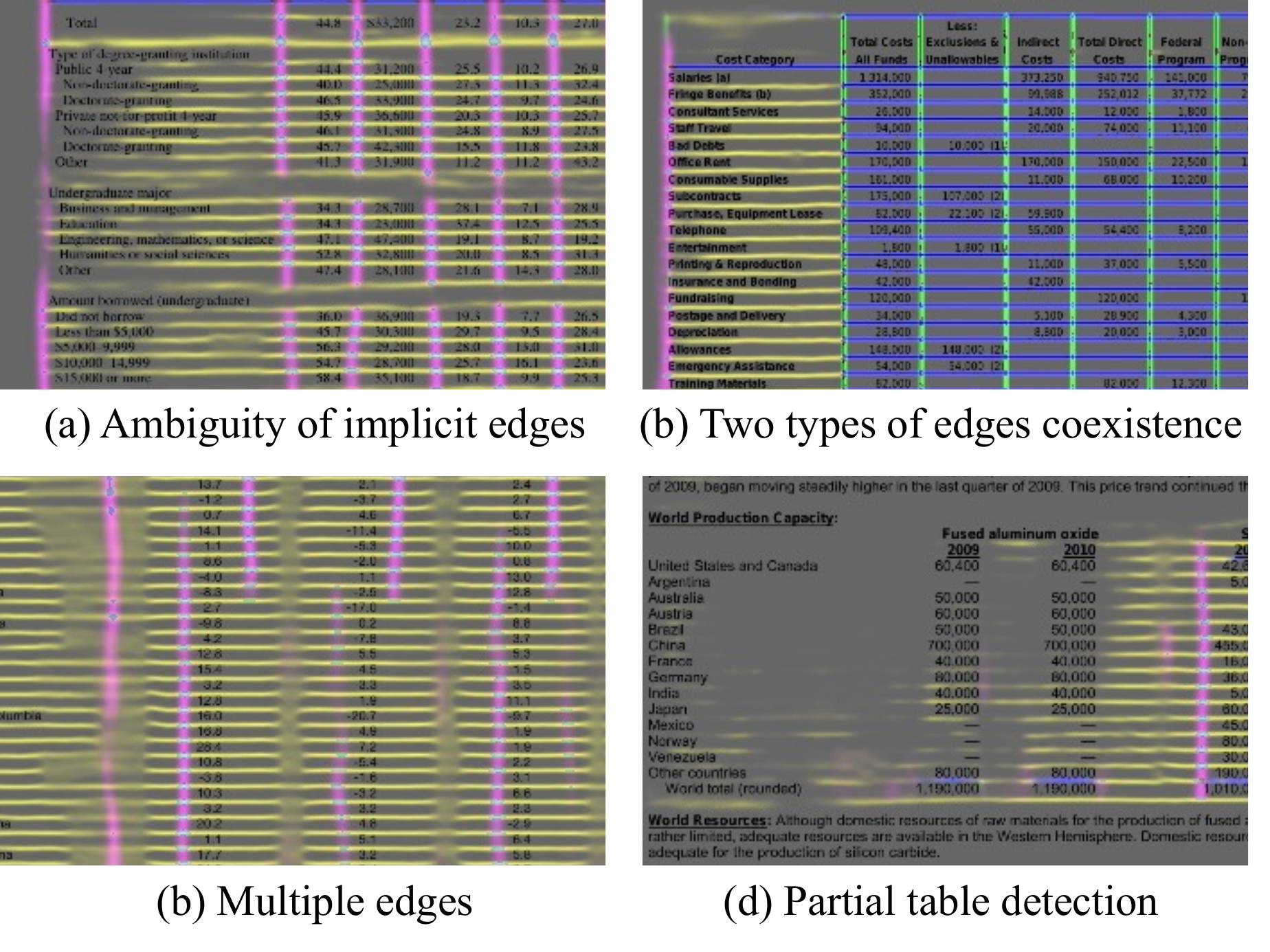}
  \caption{Failure cases caused by implicit edges in the ICDAR2013 dataset.}
  \label{fig:failurecases} 
\end{figure}

\subsubsection{Borderless table failure cases} The causes of failure on most borderless tables are analyzed in detail, as shown in Fig.~\ref{fig:failurecases}. They are roughly classified into the following four cases. (a) Two different rows merged due to implicit horizontal edges, (b) Explicit and implicit edges coexist in the same separation line, (c) Multiple vertical edges due to a wide interval between cell contents, (d) Partial table detection due to a large interval. Most of the problems arise from the ambiguity of the separation line position. To mitigate this, we need to utilize global attention-based machine learning techniques like SwinTransformer\cite{liu2021swin} as a future work.








\section{Conclusion}

We have proposed a novel end-to-end table reconstruction method, TRACE, which performs both table detection and structure recognition with a single model. This method differs from conventional approaches, which rely on a two-stage process, as it reconstructs cells and tables from fundamental visual elements such as corners and edges, in a bottom-up manner. Our model effectively recognizes tables, even when they are rotated, through the use of simple and effective post-processing techniques. We have achieved state-of-the-art performance on both the clean document dataset (ICDAR2013) and tables in the wild dataset (WTW). In future work, we plan to address weakly-supervised techniques for training the model with more diverse data.



\bibliography{main}
\bibliographystyle{splncs04}
\end{document}